\documentclass{article}

\usepackage{microtype}
\usepackage{graphicx}
\usepackage{subfigure}
\usepackage{booktabs} 

\usepackage{multirow}

\PassOptionsToPackage{hyphens}{url}
\usepackage{hyperref}

\usepackage[accepted]{icml2024}

\usepackage{amsmath}
\usepackage{amssymb}
\usepackage{mathtools}
\usepackage{amsthm}

\usepackage[capitalize,noabbrev]{cleveref}

\theoremstyle{plain}

\theoremstyle{definition}

\theoremstyle{remark}

\usepackage{algorithm}
\usepackage{algorithmic}
\usepackage{bm}
\newcommand{\sol}{{\bm{x}}}
\newcommand{\dec}{\bm{x}^\text{D}}
\newcommand{\re}{{\bm{x}^\text{R}}}
\newcommand{\solspace}{\mathcal{X}}
\newcommand{\real}{\mathbb{R}}
\newcommand{\bc}{m}
\newcommand{\bcs}{\bm{\bc}}

\newcommand{\bcsspace}{\mathcal{S}}
\newcommand{\bcsdim}{k}

\newcommand{\ar}{\mathcal{A}}

\newcommand{\maxGeneration}{T}
\newcommand{\popSize}{N}

\newcommand{\function}[1]{\texttt{#1}}

\usepackage[textsize=tiny]{todonotes}

\icmltitlerunning{Quality-Diversity with Limited Resources}

\begin{document}

\twocolumn[
\icmltitle{Quality-Diversity with Limited Resources}



\icmlsetsymbol{equal}{*}

\begin{icmlauthorlist}
\icmlauthor{Ren-Jian Wang}{nklnst,njuai}
\icmlauthor{Ke Xue}{nklnst,njuai}
\icmlauthor{Cong Guan}{nklnst,njuai}
\icmlauthor{Chao Qian}{nklnst,njuai}

\end{icmlauthorlist}

\icmlaffiliation{nklnst}{National Key Laboratory for Novel Software Technology, Nanjing University, China}
\icmlaffiliation{njuai}{School of Artificial Intelligence, Nanjing University, China}


\icmlcorrespondingauthor{Chao Qian}{qianc@nju.edu.cn}


\icmlkeywords{Quality-Diversity, Reinforcement learning, Resources-efficiency}

\vskip 0.3in
]



\printAffiliationsAndNotice{}  

\begin{abstract}
Quality-Diversity (QD) algorithms have emerged as a powerful optimization paradigm with the aim of generating a set of high-quality and diverse solutions. To achieve such a challenging goal, QD algorithms require maintaining a large archive and a large population in each iteration, which brings two main issues, sample and resource efficiency. Most advanced QD algorithms focus on improving the sample efficiency, while the resource efficiency is overlooked to some extent. Particularly, the resource overhead during the training process has not been touched yet, hindering the wider application of QD algorithms. In this paper, we highlight this important research question, i.e., how to efficiently train QD algorithms with limited resources, and propose a novel and effective method called RefQD to address it. RefQD decomposes a neural network into representation and decision parts, and shares the representation part with all decision parts in the archive to reduce the resource overhead. It also employs a series of strategies to address the mismatch issue between the old decision parts and the newly updated representation part. Experiments on different types of tasks from small to large resource consumption demonstrate the excellent performance of RefQD: it not only uses significantly fewer resources (e.g., 16\% GPU memories on QDax and 3.7\% on Atari) but also achieves comparable or better performance compared to sample-efficient QD algorithms. Our code is available at \url{https://github.com/lamda-bbo/RefQD}.
\end{abstract}

\section{Introduction}
For many real-world applications, e.g., Reinforcement Learning (RL)~\cite{DIAYN,NeuroevoSkillDiscovery}, robotics~\cite{animal,robot-few-shot-qd}, and human-AI coordination~\cite{TrajeDi,fcp}, it is required to find a set of high-quality and diverse solutions. Quality-Diversity (QD) algorithms~\citep{animal,ME,qd-survey-framework,qd-survey-optimization}, which are a subset of Evolutionary Algorithms~(EAs)~\citep{ea-book,el-book}, have emerged as a potent optimization paradigm for this challenging task. Specifically, a QD algorithm maintains a solution set (i.e., archive), and iteratively performs the following procedure: selecting a subset of parent solutions from the archive, applying variation operators (e.g., crossover and mutation) to produce offspring solutions, and finally using these offspring solutions to update the archive. The impressive performance of QD algorithms has been showcased in various RL tasks, such as exploration~\citep{first-return,trajectory-space}, robust training~\citep{one-solution,robust-training,rainbow-teaming}, and environment generation~\citep{qd4ge,dsage,warehouse}.

The goal of efficiently obtaining a set of high-quality solutions with diverse behaviors is, however, inherently difficult, posing significant challenges in the design of QD algorithms. To achieve this challenging goal, the current state-of-the-art QD algorithms require maintaining a large archive (e.g., size 1000) to save the solutions in RAM, as well as generating a large number (e.g., 100) of solutions in GPU simultaneously in each iteration. This leads to two main issues of QD algorithms to be addressed. One is the sample efficiency, i.e., how to reduce the number of samples required during the optimization of QD algorithms. Recent research has tried to improve the sample efficiency from the algorithmic perspective, by refining parent selection~\citep{qd-survey-framework,mc-me,EDO-CS,nss} and variation~\citep{ME-ES,cma-me,PGA-ME,dqd,DQD-RL,qd-pg,dcg-me} operators of QD. In these studies, it is often assumed, either explicitly or implicitly, that there is an abundant supply of computational resources. However, computational resources are typically limited in real-world scenarios, leading to the other main issue of QD, i.e., resource efficiency. Note that the performance of an algorithm depends not only on the amount of samples it used but also on its ability to handle the samples within the available computational resources. Some algorithms that can be well learned given sufficient computational resources may be poorly learned given fewer resources~\cite{theory-resources}.

The low resource efficiency of QD algorithms stems from the substantial storage and computational resources required by their execution, as we previously discussed. While there has been extensive research on sample efficiency, resource efficiency of QD has been somewhat overlooked. Recently, some studies have recognized this concern and attempted to reduce the storage overhead of the final archive during the deployment phase by condensing the archive into a single network, i.e., archive distillation~\cite{dcg-me,QDT,qd-diffusion}. However, these methods still necessitate a significant amount of storage and computational resources during the training process. As the complexity of the problems being solved increases, better (and often larger) neural networks are needed to attain satisfactory performance~\cite{rrp}, while the current QD algorithms may struggle to run under such settings. Note that the inherent difficulty of QD problems results in the high demand for storage and computational resources. Thus, \emph{how to efficiently train QD algorithms with limited resources} is a crucial yet under-explored problem that we will investigate in this study. The lack of resource-efficient QD algorithms that can effectively utilize storage and computational resources may hinder further applications of QD. It is also noteworthy that even given sufficient resources, improving the resource efficiency is still beneficial, because QD algorithms can achieve faster convergence within a limited wall-clock time, thanks to its high parallelizability~\cite{QDax}.

In this paper, we highlight, for the first time, the importance of the resource efficiency during the training process of QD algorithms, and propose a method called Resource-efficient QD (RefQD) to address this issue. Building on recent studies that disclose the distinction between the representation and decision parts in neural networks~\cite{chung2018twotimescale,dabney2021value,zhou2021over}, where the knowledge learned in the representation part can be shared and the decision part can generate diverse behaviors~\cite{ccqd}, we decompose the neural networks into these two parts, corresponding to the front layers (with numerous parameters) and the few following layers, respectively. By sharing the parameters of the representation part across all solutions, RefQD reduces the resource requirements in the training phase significantly. However, this approach introduces a tough challenge, i.e., the mismatch between the old decision parts in the archive and the newly updated representation part, resulting in the inability of the final solutions to reproduce the intended behaviors and performance. To address this issue, RefQD periodically re-evaluates the archive and weakens the elitist mechanism of QD by maintaining more decision parts in each archive cell. This allows for better utilization of the learned knowledge and reduces the wastage caused by survival selection with a frequently changing representation part. Besides, the learning rate of the representation part decays over time, ensuring more stable training and facilitating the convergence of the decision parts. 

\begin{figure}[t!]
    \centering
    \includegraphics[width=1\linewidth]{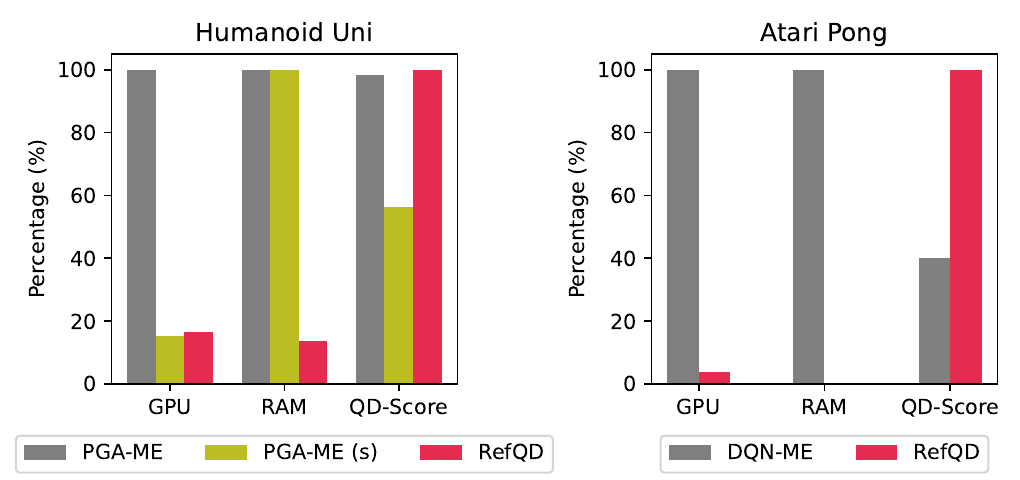}
    \caption{Performance and resource comparisons between RefQD and the baselines.}
    \vspace{-0.8em}
    \label{fig-bar}
\end{figure}

The proposed RefQD method is general, which can be implemented with different operators of advanced sample-efficient QD algorithms. In the experiments, we provide an instantiation of RefQD using uniform parent selection and variation operators from the well-known PGA-ME~\cite{PGA-ME,pga-me-telo}. We perform extensive experiments with limited resources in both the locomotion control tasks with vector-based state features~\cite{QDax} and Atari games with pixel-based image inputs~\cite{atari}. The results demonstrate that RefQD manages to reduce the GPU memory usage to 16\% and 3.7\% of the baseline in vector-based and pixel-based tasks, respectively, and reduce the RAM usage of the archive to 9.2\% and 0.3\%, respectively. Furthermore, our method still achieves the performance close to the baseline, and even better performance on some tasks. For example, Figure~\ref{fig-bar} shows the GPU, RAM, and QD-Score (the main performance metric of QD algorithms) of RefQD and the baselines on two tasks. Compared to PGA-ME (or DQN-ME), RefQD achieves a comparable (or even better) QD-Score while utilizing significantly fewer resources, demonstrating the impressive resource-efficiency of our proposed method.

Our contributions are shown as follows.
\begin{itemize}
    \item We are the first, to the best of our knowledge, to emphasize the importance of resource efficiency during the training phase of QD algorithms. 
    \item We propose RefQD, a novel and effective method to enhance the resource efficiency of QD.
    \item Experimental results with limited resources demonstrate the effectiveness of RefQD, which utilizes only 3.7\% to 16\% GPU memories and achieves comparable or even superior QD metrics, including QD-Score, Coverage, and Max Fitness, with nearly the same wall-clock time and number of samples.
\end{itemize}

\section{Background}
\subsection{Quality-Diversity}
QD algorithms~\cite{qd-survey-framework,qd-survey-optimization} aim to discover a diverse set of high-quality solutions for a given problem. Let $\solspace$ represent the solution space, and $\bcsspace \subseteq \real^\bcsdim$ denote the $\bcsdim$-dimensional behavior space of the solutions. The goal of QD algorithms is to maximize a fitness (quality) function $f: \solspace \to \real$ while exploring the $\bcsdim$-dimensional behavior space $\bcsspace$ characterized by a given behavior descriptor function $\bcs: \solspace \to \bcsspace$.
In the applications, the behavior space $\bcsspace$ can be given by an expert, trained with unsupervised learning methods~\cite{AURORA-CSC}, or learned from human feedback~\cite{divhf,qdhf}.

Taking the widely recognized QD algorithm, MAP-Elites (ME)~\cite{animal,ME}, as an example, it maintains an archive by discretizing the behavior space $\bcsspace$ into $M$ cells $\{\bcsspace_i\}_{i=1}^{M}$ and storing at most one solution in each cell.
The goal of ME is formalized as maximizing
the sum $\sum\nolimits_{i=1}^M f(\sol_i)$ of the fitness of the solutions in each cell, i.e., the QD-Score,
where $\sol_i$ denotes the solution contained within the cell $\bcsspace_i$, i.e., $\bcs(\sol_i) \in \bcsspace_i$.
If a cell $\bcsspace_i$ does not contain a solution, then the corresponding fitness value $f(\sol_i)$ is considered as $0$. For simplicity, the fitness value $f(\cdot)$ is assumed (or transformed) to be non-negative to prevent the QD-Score from decreasing.
The main process of ME consists of selecting parent solutions from the archive, generating offspring solutions through variation operators, evaluating the offspring solutions, and updating the archive (i.e., survivor selection).
As the goal is to fill the cells with high-quality solutions, ME saves the solution with the best fitness in each cell in survivor selection.

Due to the inherent difficulty of obtaining a set of high-quality solutions with diverse behaviors, QD algorithms often require a large number of samples during optimization. Thus, how to improve the sample efficiency of QD has become a critical problem.
Based on the improved components of QD~\cite{qd-survey-framework}, recent studies can be roughly divided into two categories: how to select parent solutions from the archive (i.e., parent selection)~\citep{NSLC,qd-survey-framework,EDO-CS,nss}, and how to update them to reproduce offspring solutions (i.e., variation)~\cite{ME-ES,cma-me,PGA-ME,dqd,qd-pg,pga-me-telo}. Apart from sample efficiency, the resource efficiency (e.g., GPU memory overhead during the training phase) is also a big issue of QD algorithms, which we will discuss in this work.

\subsection{Archive Distillation}
Some recent works have noticed the resources overhead of the QD algorithms in the deployment phase, and manage to reduce the resources of the final archive by archive distillation.
Distilling the knowledge of an archive into a single policy is an alluring process that reduces the number of parameters output by the algorithm and enables generalization. 
\citet{dcg-me} introduced the descriptor-conditioned gradient and distilled the experience of the archive into a single descriptor-conditioned actor.   
\citet{QDT} distilled the knowledge of the archive into a single decision transformer.
\citet{qd-diffusion} used diffusion models to distill the archive into a single generative model over policy parameters. These methods can achieve a high compression ratio while
recovering most QD metrics. However, it is still required to maintain a large archive as well as a large number of solutions \emph{during the training process}.
Our proposed RefQD reduces the resource overhead in both the training and deployment phases. In addition, RefQD can also be combined with the archive distillation methods to obtain a single network. As an example, the deep decision archive of RefQD can be distilled to further reduce the resource overhead in the deployment phase.

\section{Method}
In this section, we will introduce the proposed RefQD method. RefQD follows the common procedure of QD algorithms, i.e., iterative parent selection, offspring solution reproduction, and archive updating. Inspired by the observation that the front and following layers of a neural network are used for state representation and decision-making, respectively~\cite{zhou2021over,ccqd}, RefQD decomposes a neural network into a representation part and a decision part (e.g., containing only the last layer of the neural network), and stores only decision parts in the archive while sharing a unique representation part, which will be introduced in Section~\ref{sec-decomp-share}. Such a strategy can improve the resource efficiency significantly, but also leads to the mismatch issue, i.e., a decision part in the archive fails to reproduce its behavior and fitness when combined with the latest shared representation part due to the frequent change of the representation part. Section~\ref{sec-overcoming-mismatch} then introduces a series of strategies to overcome the mismatch issue, including periodic re-evaluation, maintaining a deep decision archive, top-$k$ re-evaluation, and learning rate decay for the representation part. In Section~\ref{sec-refqd}, we will present the RefQD method in detail along with its pseudo-codes in Algorithm~\ref{alg:refqd}. 

\subsection{Decomposition and Sharing}\label{sec-decomp-share}

The low resource efficiency of QD algorithms is mainly due to the maintenance of too many solutions. However, in order to achieve sufficient diversity, it is unavoidable to maintain and update many solutions simultaneously. Therefore, to improve the resource efficiency, we try to reduce the total number of parameters of these solutions instead.

Considering the recent observation that different layers of a neural network have different functions~\citep{chung2018twotimescale,dabney2021value,zhou2021over,re2,ccqd}, we divide a network into two parts, where the front layers with numerous parameters are considered as the representation part, and the few following layers are considered as the decision part. The knowledge learned by the representation part is usually common and can be shared, while the diverse behaviors can be generated by different decision parts. To significantly reduce the resource overhead, we maintain \emph{only one} representation part, sharing it across all the decision parts in the archive and offspring population. In this way, we can reduce the RAM overhead from the archive and the GPU memory overhead from the offspring population simultaneously. This allows QD algorithms to be applied to larger scale problems with limited resources.

\subsection{Overcoming Mismatch Issue}\label{sec-overcoming-mismatch}
\textbf{Mismatch issue.} In the QD algorithms, it is expected for the solutions in each cell of the archive to reproduce the corresponding behaviors and fitness when they are selected as parent solutions or used in the deployment phase.
Note that many decision parts in the archive can only reproduce the corresponding behaviors and fitness when combined with the corresponding versions of the representation parts at the time they were added to the archive. 
However, as we only maintain one representation part and do not save the representation parts of the solutions along with their decision parts in the archive, we can only use the latest version of the shared representation part to be combined with the decision parts.
Since the latest representation part is being updated frequently during the iteration process, the decision parts in the archive may fail to reproduce their behaviors and fitness when combined with the latest shared representation part.
Even worse, when a new decision part matching the latest version of the representation part attempts to add to the corresponding cell in the archive, it is likely to be eliminated by the decision part already in the cell with a better yet outdated performance record that it can no longer reproduce. 
We refer to this as the \emph{mismatch} issue between the decision parts in the archive and the latest shared representation part.

\textbf{Periodic re-evaluation.} To address the mismatch issue, it is natural to periodically re-evaluate all the decision parts in the archive with the latest shared representation part, and add them back to the archive by survivor selection. However, we have observed that as the representation part changes frequently, there are a large number of ``dead" decision parts in each time of re-evaluation, which have low fitness led by the mismatch issue and will be deleted in the survivor selection procedure. This will waste a lot of learned knowledge in the decision parts. In fact, many of the decision parts that do not match the current version of representation part can match a later one. Thus, only using periodic re-evaluation cannot address the mismatch issue.

\textbf{Deep Decision Archive (DDA).} To alleviate the wastage of learned knowledge led by periodic re-evaluation, RefQD weakens the elitist mechanism of QD by maintaining $K$ decision parts (e.g., $K=4$ in our experiments) instead of only one in each cell of the archive. That is, RefQD now maintains a DDA $\mathcal A = \{\mathcal A^{(i)}\}^K_{i=1}$ with $K$ levels, where each level of a cell in the archive stores one solution, and the fitness decreases as the level increases. Thus, the first level of DDA corresponds to the decision archive of traditional QD algorithms. Note that using a DDA brings a more robust exploration, because an inferior decision part when combined with the current representation part may become better when combined with later representation parts. When updating DDA, if the solution in the $i$-th level of a cell is replaced by survivor selection, it will not be removed directly, while the solution in the $K$-th level (i.e., the worst solution in the current cell) will be removed, and the solutions contained by the $i$-th to $(K-1)$-th level will move down one level accordingly. The idea of deep archive has also been used to solve uncentainty in noisy QD domains~\cite{adaptive-sampling,DG-ME} and noisy general EAs~\cite{porss,dposs}. Note that the decision part is much smaller than the representation part. Thus, maintaining a DDA (i.e., storing multiple decision parts in each cell of the archive) will not significantly affect the resource efficiency.

\textbf{Top-$k$ re-evaluation.} Although using a DDA can reduce the wastage of learned knowledge, re-evaluating the entire DDA will need many samples and cost much time. To make a trade-off, we propose top-$k$ re-evaluation to re-evaluate only the solutions in the top $k$ levels of the DDA, as our final goal is to obtain a diverse set of high-quality solutions.

\textbf{Learning rate decay for the representation part.} The representation part changes frequently during the optimization of QD, making the decision parts hard to converge. Thus, we decay the learning rate of the representation part with iterations to improve the stability and allow the decision parts a better convergence.

\begin{algorithm}[t]
\caption{Resource-efficient QD}
\label{alg:refqd}
\textbf{Input}: Number $\maxGeneration$ of total generations, number $\popSize$ of selected solutions in each generation, number $M$ of cells of each level of DDA, period $T_r$ for re-evaluation, number $k$ of top levels of the archive to be re-evaluated\\
\textbf{Output}: Representation part $\re$, first level $\ar^{(1)}$ of DDA \\\vspace{-1.2em}
\begin{algorithmic}[1]
    \STATE Let $\ar \gets \emptyset$, $t \gets 1$;
    \WHILE {$t \leq \maxGeneration$}
        \IF {$t = 1$}
            \STATE $\re, \{\dec_{i}\}_{i=1}^{N} \gets \function{Randomly\_Generate}(\popSize)$
        \ELSE
            \STATE $\re \gets \function{Train\_Representation}(\re, \ar)$;
            \STATE $\{\dec_{i}\}_{i=1}^{N} \gets \function{Parent\_Selection}(\ar, \popSize)$;
            \STATE $\{{\dec_{i}}'\}_{i=1}^{N} \gets  \function{Variation}(\re, \{\dec_{i}\}_{i=1}^{N})$
        \ENDIF
        \STATE $\function{Evaluation}(\re, \{{\dec_{i}}'\}_{i=1}^{N})$;
        \STATE $\ar \gets \function{DDA\_Selection}(\ar, \{{\dec_{i}}'\}_{i=1}^{N})$;
        \STATE $t \gets t+1$;
        \IF {$t\!\! \mod T_r = 0$}
            \STATE $\ar', \{\dec_{i}\}_{i=1}^{M \times k} \gets \function{Take\_Levels}(\ar, k)$;
            \STATE $\function{Re\_Evaluation}(\re, \{\dec_{i}\}_{i=1}^{M \times k})$;
            \STATE $\ar \gets \function{DDA\_Selection}(\ar', \{\dec_{i}\}_{i=1}^{M \times k})$
        \ENDIF
    \ENDWHILE
    \STATE \textbf{return} $\re$, $\ar^{(1)}$
\end{algorithmic}
\end{algorithm}

\begin{algorithm}[htb]
\caption{DDA Selection}
\label{alg:dda}
\textbf{Input}: DDA $\mathcal A = \{\mathcal A^{(i)}\}^K_{i=1}$ with $K$ levels, decision parts $\{\dec_{i}\}_{i=1}^n$ to be added\\
\textbf{Output}: Updated DDA $\mathcal A = \{\mathcal A^{(i)}\}^K_{i=1}$ with $K$ levels\\\vspace{-1.2em}
\begin{algorithmic}[1]
    \FOR {$i = 1$ \textbf{to} $n$}
        \STATE $j \gets \function{Get\_Cell\_Index}(\bcs(\dec_i))$;
        \STATE $l \gets 1$;
        \WHILE {$l \le K$}
            \IF {$\ar^{(l)}_j = \emptyset$}
                \STATE $\ar^{(l)}_j \gets \boldsymbol \dec_i$;
                \STATE \textbf{break}
            \ELSIF{$f(\ar^{(l)}_j) < f(\dec_i)$}
                \STATE $\bm{z} \gets \ar^{(l)}_j$, $\ar^{(l)}_j \gets \dec_i$, $\dec_i \gets \bm{z}$

            \ENDIF
            \STATE $l \gets l + 1$
        \ENDWHILE
    \ENDFOR
    \STATE \textbf{return} $\ar$
\end{algorithmic}
\end{algorithm}

\subsection{Resource-efficient QD}\label{sec-refqd}
The procedure of RefQD is described in Algorithm~\ref{alg:refqd}. 
At the beginning, the DDA $\ar$ is created as an empty set in line~1, and a shared representation part $\re$ and $N$ decision parts $\{\dec_{i}\}_{i=1}^{N}$ are randomly generated in line~4. After that, in each generation $t$ (where $1 < t \leq T$), RefQD first trains the representation part $\re$ in line~6, then selects $N$ decision parts $\{\dec_{i}\}_{i=1}^{N}$ from the DDA $\ar$ as parents in line~7, and finally combines each of them with the representation part $\re$ to form $N$ complete policies which are conducted by variation to generate $N$ offspring solutions $\{{\dec_{i}}'\}_{i=1}^{N}$ in line~8. After evaluating the generated offspring solutions $\{{\dec_{i}}'\}_{i=1}^{N}$ in line~10, we get the fitness and behavior of each solution. 
Then, we use these offspring solutions to update the DDA $\ar$ in line~11, by employing Algorithm~\ref{alg:dda}. After every $T_r$ iterations, top-$k$ re-evaluation is applied and the DDA $\ar$ is updated accordingly, as shown in lines~14--16 of Algorithm~\ref{alg:refqd}. That is, the solutions $\{\dec_{i}\}_{i=1}^{M \times k}$ in the top $k$ levels of the DDA $\ar$ are re-evaluated by combining with the current representation part $\re$, and then used to update the DDA $\ar'$ which contains the remaining solutions after deleting $\{\dec_{i}\}_{i=1}^{M \times k}$ from $\ar$.

Algorithm~\ref{alg:dda} presents the procedure of using a set of decision parts (or called solutions) $\{\dec_{i}\}_{i=1}^n$ to update a DDA $\mathcal A = \{\mathcal A^{(i)}\}^K_{i=1}$ with $K$ levels. Each solution $\dec_{i}$ will be compared with the solutions archived in the corresponding cell (indexed by $j$ in line~2) of the DDA, from the top level to the bottom level (i.e., lines~4--12). If the current level $\ar^{(l)}_j$ is empty, i.e., $\ar^{(l)}_j = \emptyset$ in line~5, it will be occupied by $\dec_{i}$ directly in line~6. If it is not empty and the contained solution is worse than $\dec_{i}$ (i.e., $f(\ar^{(l)}_j)<f(\dec_{i})$ in line~8), the current level $\ar^{(l)}_j$ will also be occupied by $\dec_{i}$, as $\ar^{(l)}_j \gets \dec_i$ shown in line~9. In the latter case, the old solution in any level $\ar^{(p)}_j$ (where $p \geq l$) will be moved to the following level $\ar^{(p+1)}_{j}$, which is accomplished by using $\dec_{i}$ to record the old solution in $\ar^{(l)}_j$ (i.e., $\bm{z} \gets \ar^{(l)}_j$, $\dec_i \gets \bm{z}$ in line~9) and continuing the loop. Thus, if all the $K$ levels of cell $j$ have solutions, the solution in the last level $\ar^{(K)}_j$, i.e., the worst solution, will be actually removed.

Note that the proposed RefQD is a general framework that can be implemented with different components. In our experiment, we will provide an implementation using the uniform parent selection and reproduction operators of PGA-ME~\cite{PGA-ME,pga-me-telo}.

\section{Experiment}

\subsection{Experimental Settings}

To examine the performance of RefQD, we conduct experiments on the QDax suite and Atari environments.

\textbf{QDax\footnote{\url{https://github.com/adaptive-intelligent-robotics/QDax}}.} It is a popular framework and benchmark for QD algorithms~\cite{QDax}. We conduct experiments on five unidirectional tasks and three path-finding tasks. The unidirectional tasks aim to obtain a set of robot policies that run as fast as possible with different frequency of the usage of the feet, including Hopper Uni, Walker2D Uni, HalfCheetah Uni, Humanoid Uni, and Ant Uni. The reward is mainly determined by the forward speed of the robot, and the behavior descriptor is defined by the fraction of time each foot contacting with the ground. The path-finding tasks aim to obtain a set of robots that can reach each location on the given map, including Point Maze, Ant Maze, and Ant Trap. The behavior descriptor is defined as the final position of the robot. The reward is defined as the negative distance to the target position in Point Maze and Ant Maze, and the distance forward in Ant Trap. In all the tasks, the observation is a vector of sensor data, and the action is a vector representing the torque of each effector.

\textbf{Atari.} To further investigate the versatile of RefQD, we also conduct experiments on the video game Atari~\cite{atari}, which is a widely used benchmark in RL. The observation of Atari is an $84 \times 84$ image of the video game, and the action is the discrete button to be pressed. 
We conduct experiments on two tasks, i.e., Pong and Boxing. For Pong, the one-dimensional behavior descriptor is defined as the frequency of the movement of the agent. For Boxing, the two-dimensional behavior descriptor is defined as the frequency of the movement and the frequency of the punches of the agent, respectively.
The agent will receive positive and negative rewards when it wins or loses points in the games, respectively.
Compared with QDax, Atari has a large image-based observation space and typically requires a larger CNN or ResNet~\cite{resnet} to process visual data, posing more challenges for the QD algorithms.

\begin{figure*}[t!]\centering
\includegraphics[width=0.96\linewidth]{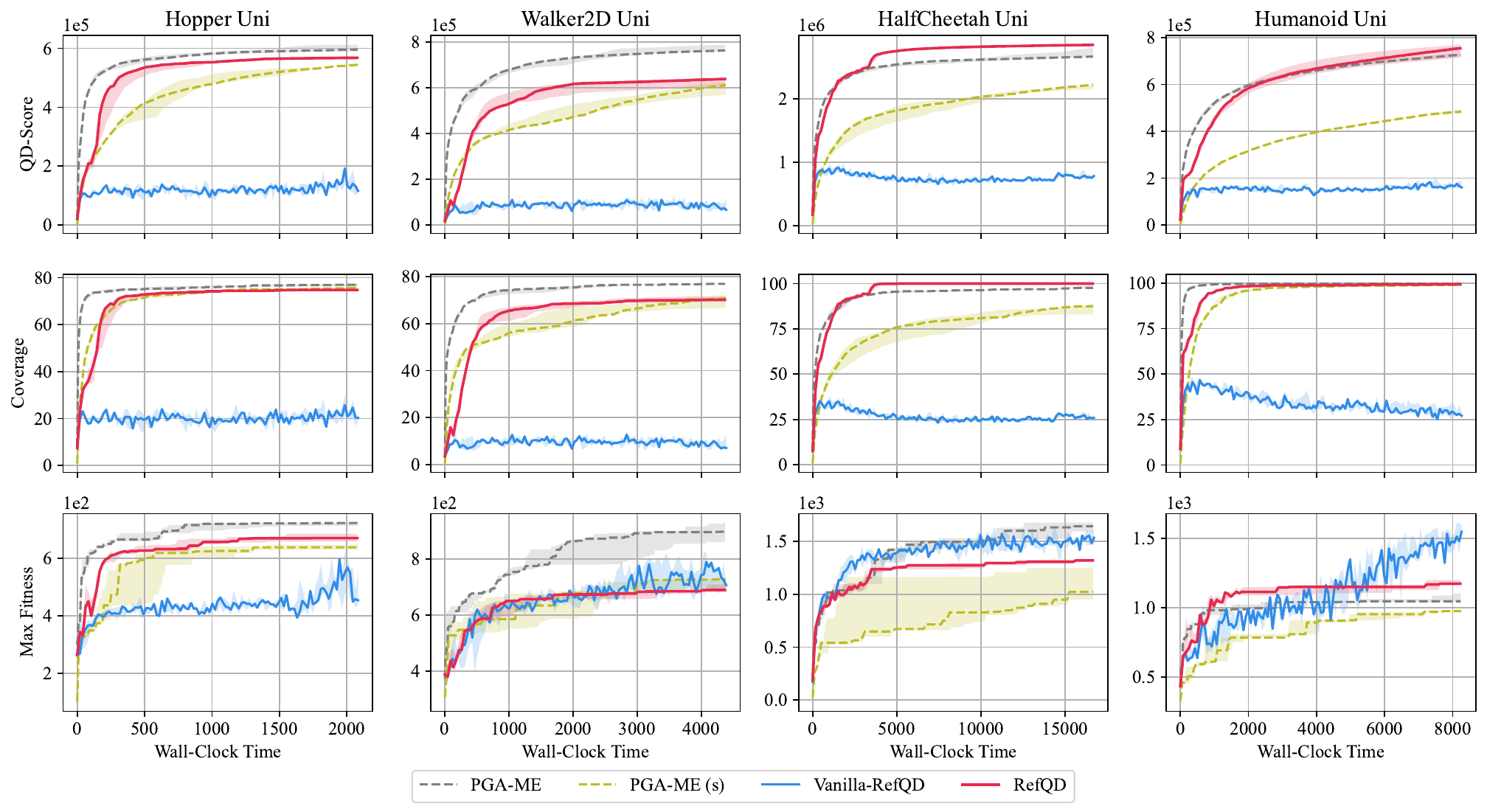}
\includegraphics[width=0.96\linewidth]{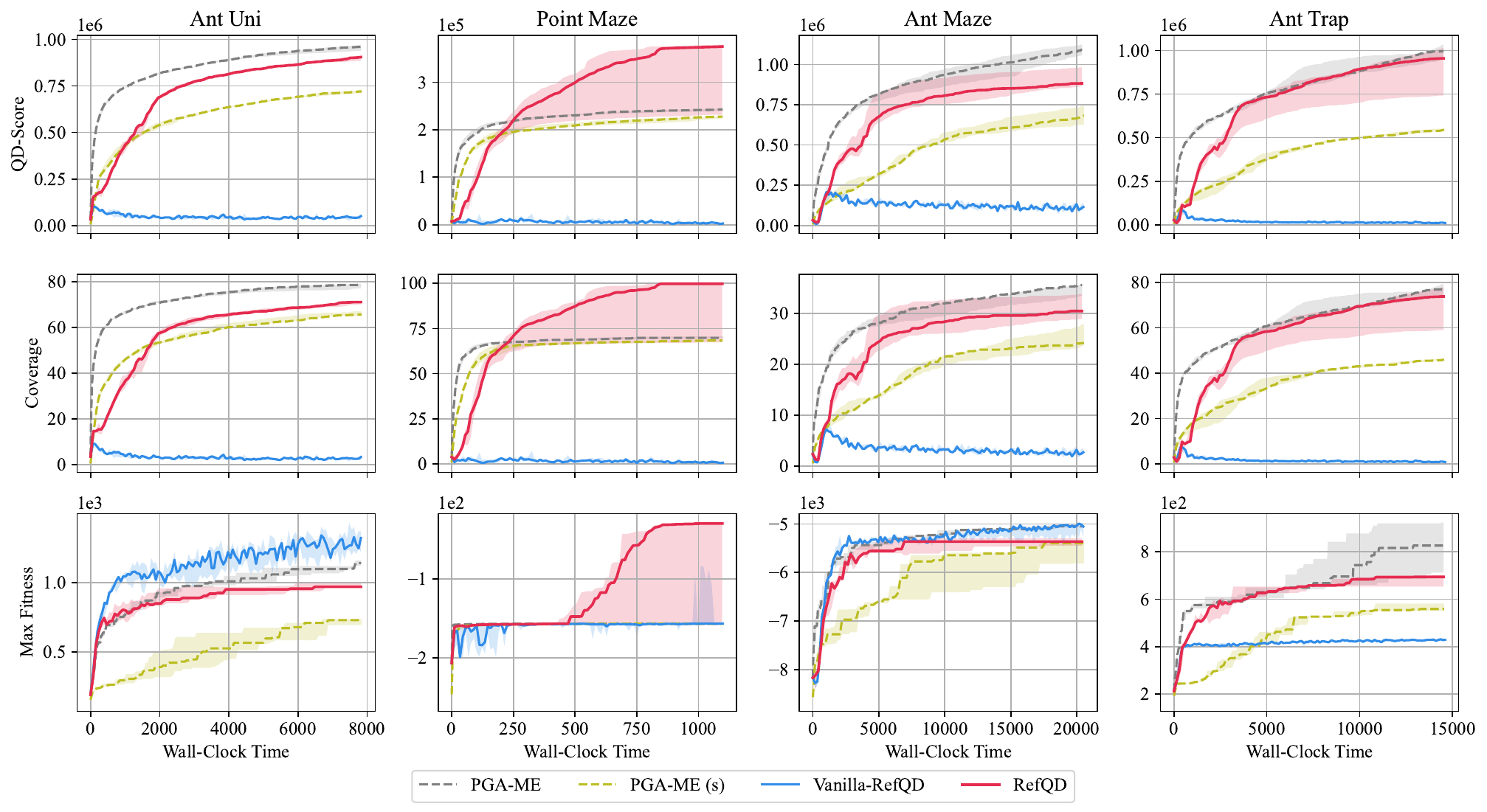}
\caption{Performance comparison in terms of QD-Score, Coverage, and Max Fitness on eight environments of QDax. The medians and the first and third quartile intervals are depicted with curves and shaded areas, respectively.}\label{fig-QDax}
\end{figure*}

To evaluate the effectiveness of RefQD, we compare the following methods.
\begin{itemize}
    \item \textbf{RefQD}: Our proposed method.
    \item \textbf{Vanilla-RefQD}: The vanilla version of RefQD, which only uses decomposition and sharing, and does not use the strategies for overcoming the mismatch issue.
    \item \textbf{PGA-ME} or \textbf{DQN-ME}: PGA-ME / DQN-ME with the \emph{default settings}. DQN-ME is the method that replaces TD3~\cite{td3} with Double DQN~\cite{dqn,ddqn} in PGA-ME for the tasks with discrete action spaces on Atari.
    \item \textbf{PGA-ME (s)} or \textbf{DQN-ME (s)}: PGA-ME / DQN-ME with a small number of offspring solutions, which have a much smaller GPU memory overhead compared to the default settings. However, they still maintain the whole archive (including many large representation parts), which costs lots of RAM, while all the RefQD variants do not need to.
\end{itemize}

We consider the following QD metrics: 
1) \textbf{QD-Score}: The total sum of the fitness values across all solutions in the archive. It measures both the quality and the diversity of the solutions, which is the most important metric for the QD algorithms.
2) \textbf{Coverage}: The percentage of cells that have been covered by the solutions in the archive. It can measure the diversity of the solutions.
3) \textbf{Max Fitness}: The highest fitness value of the solutions in the archive. It can measure the quality of the solutions.

For a fair comparison, all the RefQD variants use the same reproduction operator with the compared method, i.e., same to PGA-ME on QDax and same to DQN-ME on Atari.
All the RefQD variants use a decomposition strategy ($n+1$), i.e., use the last layer as the decision part and the rest as the representation part.
In order to compare the performance under limited resources, all the methods are \textbf{given the same amount of GPU memory}, except for PGA-ME and DQN-ME, which are given unlimited GPU memory.
In addition, it is \textbf{not allowed to maintain the whole archive} (i.e., to have cells that contain both representation and decision networks) for all methods,
except for PGA-ME (DQN-ME) and PGA-ME (DQN-ME) (s), which have to store a full policy in each cell of the archive.
Detailed settings can be found in Appendix~\ref{app:detailed-settings} due to space limitation. Our code is available at \url{https://github.com/lamda-bbo/RefQD}.

\subsection{Main Results}

We plot the QD-Score, Coverage, and Max Fitness curves for different methods. 
Figure~\ref{fig-QDax} and Table~\ref{table:resource-mlp} show the results and the resource usage on eight environments of QDax, respectively.
We can observe that RefQD with the limited resources, achieves the similar performance to PGA-ME that uses unlimited resources, and performs even better in HalfCheetah Uni, Humanoid Uni, and Point Maze. 
RefQD generally uses much less GPU memories, e.g., only using 16\% GPU memories on Ant Uni compared to PGA-ME.
PGA-ME (s) becomes extremely slow, performing significantly worse than RefQD and PGA-ME, although it still maintains the whole archive that costs lots of RAM. 
PGA-ME demonstrates the best overall QD-Score. However, it utilizes a substantial amount of GPU memories as well as a significant amount of RAM, as demonstrated in Figure~\ref{fig-bar}. As is stated in Section~\ref{sec-overcoming-mismatch}, Vanilla-RefQD has the serious mismatch issue, resulting in the worst performance here.

\begin{table*}[htbp]
\caption{Resource usage and QD-Score comparisons of different methods on eight environments of QDax.}
\vspace{0.3em}
\centering
\begin{tabular}{c|ccc|ccc|ccc} \toprule
\multirow{2}*{Tasks} & \multicolumn{3}{c|}{PGA-ME} & \multicolumn{3}{c|}{PGA-ME (s)} & \multicolumn{3}{c}{RefQD} \\
& GPU & RAM & QD-Score & GPU & RAM & QD-Score & GPU & RAM & QD-Score \\ \midrule
Hopper Uni      & $100.0\%$ & $100.0\%$ & $100.0\%$ & $13.3\%$ & $100.0\%$ & $76.5\%$ & $14.2\%$ & $4.5\%$ & $92.7\%$ \\
Walker2D Uni    & $100.0\%$ & $100.0\%$ & $100.0\%$ & $13.5\%$ & $100.0\%$ & $63.9\%$ & $14.9\%$ & $8.7\%$ & $79.6\%$ \\
HalfCheetah Uni & $100.0\%$ & $100.0\%$ & $100.0\%$ & $13.5\%$ & $100.0\%$ & $74.0\%$ & $14.9\%$ & $8.6\%$ & $104.6\%$ \\
Humanoid Uni    & $100.0\%$ & $100.0\%$ & $100.0\%$ & $15.2\%$ & $100.0\%$ & $57.1\%$ & $16.5\%$ & $13.7\%$ & $101.6\%$ \\
Ant Uni         & $100.0\%$ & $100.0\%$ & $100.0\%$ & $14.3\%$ & $100.0\%$ & $67.8\%$ & $16.0\%$ & $9.2\%$ & $84.4\%$ \\
Point Maze      & $100.0\%$ & $100.0\%$ & $100.0\%$ & $13.3\%$ & $100.0\%$ & $88.7\%$ & $14.0\%$ & $3.2\%$ & $85.0\%$ \\
Ant Maze        & $100.0\%$ & $100.0\%$ & $100.0\%$ & $14.3\%$ & $100.0\%$ & $31.4\%$ & $15.6\%$ & $8.9\%$ & $64.9\%$ \\
Ant Trap        & $100.0\%$ & $100.0\%$ & $100.0\%$ & $14.3\%$ & $100.0\%$ & $57.4\%$ & $15.6\%$ & $9.0\%$ & $107.1\%$ \\ \bottomrule
\end{tabular}
\label{table:resource-mlp}
\end{table*}

\begin{figure}[htbp]\centering
\includegraphics[width=0.96\linewidth]{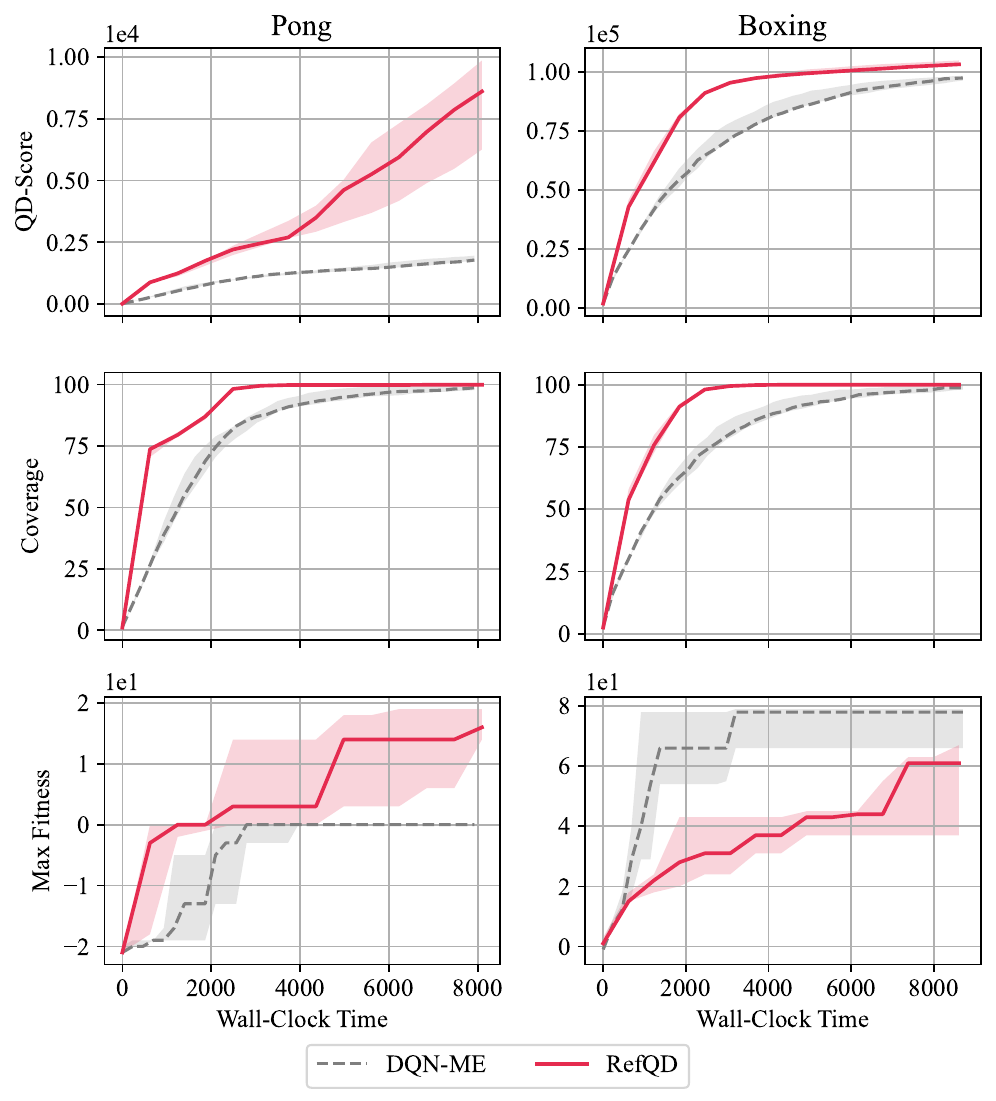}
\caption{Performance comparison in terms of QD-Score, Coverage, and Max Fitness using CNN as policy networks on two environments of Atari. The medians and the first and third quartile intervals are depicted with curves and shaded areas, respectively.}\label{fig-Atari}
\end{figure}

The experimental results and the resource usage on two Atari environments with CNN policy architecture are shown in Figure~\ref{fig-Atari} and Table~\ref{table:resource-cnn}, respectively. Due to the complex observation space of Atari, DQN-ME and RefQD have a much larger representation part on this task, leading to a more obvious advantage of RefQD on the resource efficiency.
Besides, the QD-Scores of RefQD on both tasks are also much better than those of DQN-ME. This is because the employed network decomposition can also simplify the optimization space of QD, thus making it easier to find better solutions, as also observed in CCQD~\cite{ccqd}.

We also conduct experiments on the Atari environments with ResNet policy architecture to examine the scalability of RefQD in challenging tasks with higher-dimensional decision space.
The default settings of DQN-ME run out of memory and fail to work, while RefQD with limited resource performs significantly better than DQN-ME (s) with unlimited RAM, as shown in Table~\ref{table:resource-resnet}, demostrating that RefQD makes it possible to solve larger-scale QD problems efficiently.
Due to space limitation,
detailed results of QD metrics can be found in Appendix~\ref{app:resnet}.

\begin{table}[htbp]
\caption{Resource usage and QD-Score comparisons of different methods using CNN as policy networks on two environments of Atari.}
\vspace{0.3em}
\centering
\resizebox{1.0\linewidth}{!}{
\begin{tabular}{c|ccc|ccc} \toprule
\multirow{2}*{Tasks} & \multicolumn{3}{c|}{DQN-ME} & \multicolumn{3}{c}{RefQD} \\
& GPU & RAM & QD-Score & GPU & RAM & QD-Score \\ \midrule
Pong & $100.0\%$ & $100.0\%$ & $100.0\%$ & $3.7\%$ & $0.3\%$ & $248.6\%$ \\
Boxing & $100.0\%$ & $100.0\%$ & $100.0\%$ & $3.7\%$ & $0.6\%$ & $119.6\%$ \\ \bottomrule
\end{tabular}
}
\label{table:resource-cnn}
\end{table}

\begin{table}[htbp]
\caption{Resource usage and QD-Score comparisons of different methods using ResNet as policy networks on two environments of Atari.}
\vspace{0.3em}
\centering
\resizebox{1.0\linewidth}{!}{
\begin{tabular}{c|ccc|ccc} \toprule
\multirow{2}*{Tasks} & \multicolumn{3}{c|}{DQN-ME (s)} & \multicolumn{3}{c}{RefQD} \\
& GPU & RAM & QD-Score & GPU & RAM & QD-Score \\ \midrule
Pong & $100.0\%$ & $100.0\%$ & $100.0\%$ & $52.5\%$ & $2.5\%$ & $220.1\%$ \\
Boxing & $100.0\%$ & $100.0\%$ & $100.0\%$ & $52.5\%$ & $2.6\%$ & $246.2\%$ \\\bottomrule
\end{tabular}
}
\label{table:resource-resnet}
\end{table}

\subsection{Additional Results}
We conduct additional experiments to deeply analyze the proposed RefQD, by investigating the influence of the decomposition strategy, the number $K$ of levels of DDA, the period $T_r$ of re-evaluations, and the number $k$ of top-$k$ re-evaluations. Other experiments, including ablation studies of RefQD, performance analysis based on time-steps, and experiments based on another state-of-the-art QD algorithm EDO-CS~\cite{EDO-CS}, are provided in Appendix~\ref{app:additional-results} due to space limitation.

\textbf{Influence of the decomposition strategy.} Generally, the more we share, the less computational resources are required, but the greater potential for performance drop. We conduct experiments to analyze the influence of the decomposition strategies. When sharing less representation, RefQD has a better Max Fitness, due to the stronger ability from additional resource overhead, as shown in Appendix~\ref{app:decomp}. In practical use, the decomposition strategy can be set according to the available computational resources.

\textbf{Influence of the number $K$ of levels of DDA.} We analyze the influence of the number $K$ of levels of DDA, where $K$ is set to 1, 2, and 4 on all the eight tasks of QDax. The large $k$ DDA has, the more history knowledge will be saved in it, which also costs more RAM usage. As shown in Table~\ref{table:depth}, using $K=4$ achieves the best QD-Score on most tasks except for Ant Maze and Ant Trap. Larger $K$ may bring better QD-Score but also leads to the usage of more resources. Thus, we choose $4$ as the default value used in our experiments.

\begin{table}[t!]
\caption{QD-Score of RefQD by using different number $K$ of levels of DDA on eight environments of QDax.}\vspace{0.3em}
\centering
\begin{tabular}{c|ccc} \toprule
Tasks & $K=1$ & $K=2$ & $K=4$ \\ \midrule
Hopper Uni & $472169$ & $491220$ & $\textbf{500594}$ \\
Walker2D Uni & $423790$ & $466840$ & $\textbf{566845}$ \\
HalfCheetah Uni & $2476923$ & $2462211$ & $\textbf{2629326}$ \\
Humanoid Uni & $587483$ & $632468$ & $\textbf{646501}$ \\
Ant Uni & $490440$ & $678343$ & $\textbf{715349}$ \\
Point Maze & $188707$ & $190907$ & $\textbf{192868}$ \\
Ant Maze & $620880$ & $\textbf{738880}$ & $669850$ \\
Ant Trap & $\textbf{919987}$ & $752117$ & $741243$ \\
\bottomrule
\end{tabular}
\label{table:depth}
\end{table}

\begin{figure}[t!]\centering
\includegraphics[width=1\linewidth]{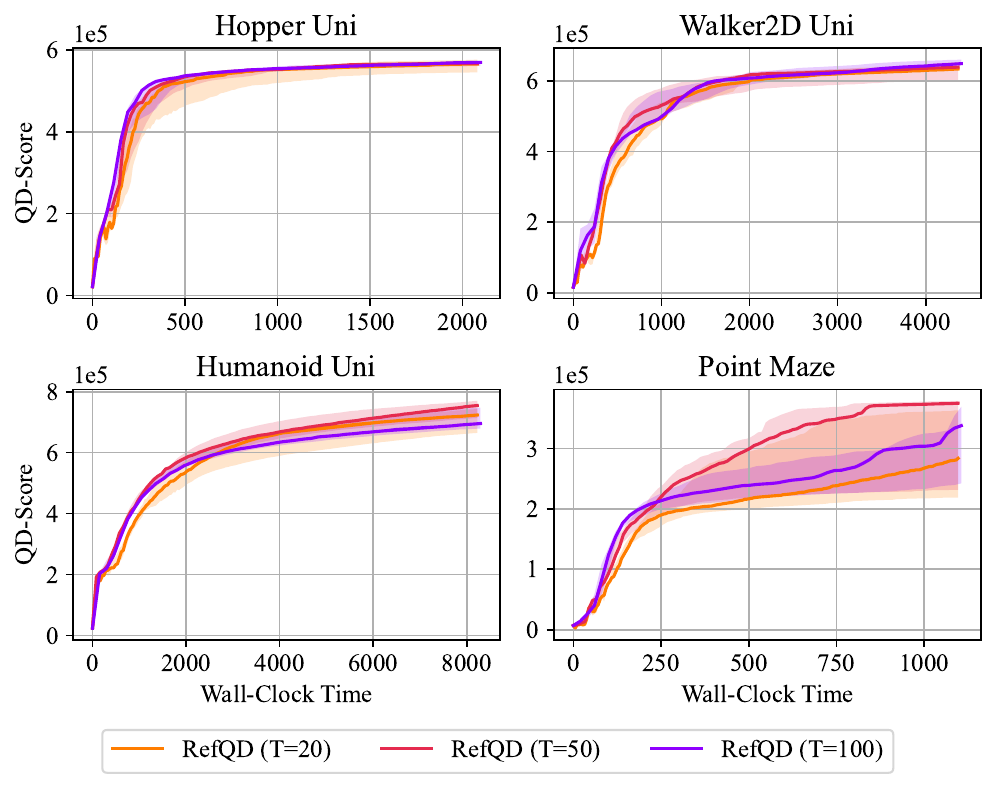} 
\vspace{-0.8em}
\caption{QD-Score of RefQD with different period $T_r$ of re-evaluation on four environments of QDax.}\label{fig-reeval-period}
\end{figure}

\textbf{Influence of the period $T_r$ of re-evaluations.} Then, we perform the sensitivity analysis on the hyperparameter $T_r$. We compare three values of $T_r$, i.e., 20, 50, and 100. As shown in Figure~\ref{fig-reeval-period}, their performances on most environments are similar, but $T_r=50$ generally performs better. Thus, we use 50 as the default value for $T_r$.

\begin{figure}[t!]\centering
\includegraphics[width=1\linewidth]{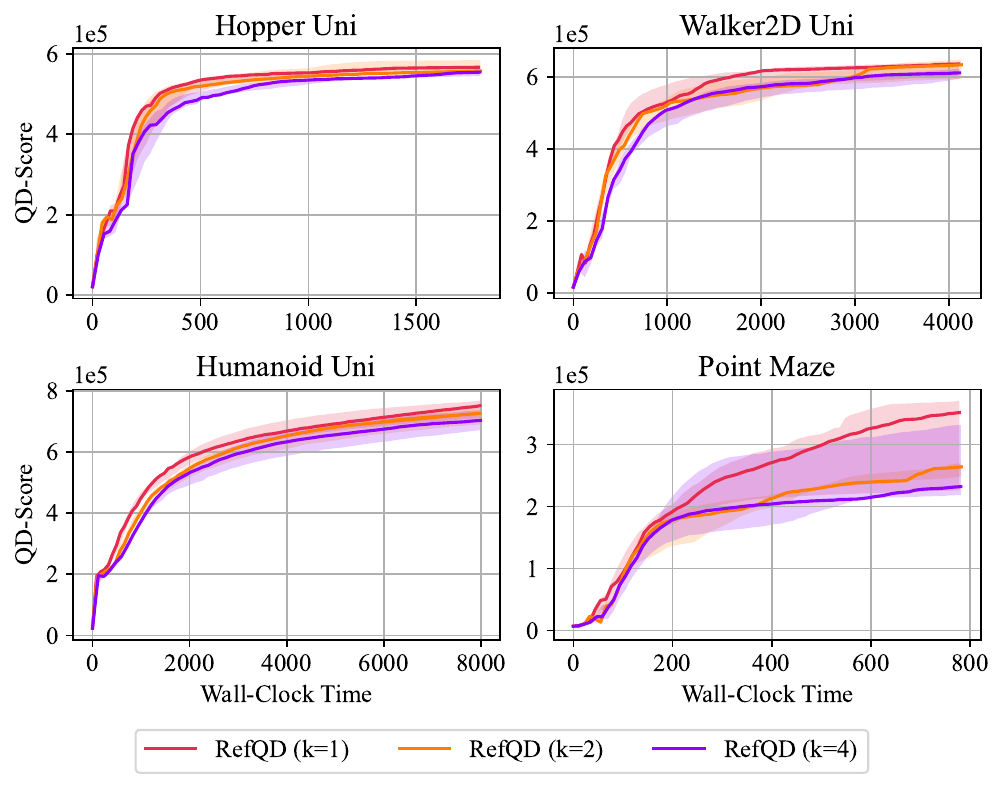} 
\vspace{-0.8em}
\caption{QD-Score of RefQD with different number $k$ of top-$k$ re-evaluation on four environments of QDax.}\label{fig-top-k-reeval}
\end{figure}

\textbf{Influence of the number $k$ of top-$k$ re-evaluation.} The top-$k$ re-evaluation strategy is proposed to make a trade-off between the number of re-evaluated samples and the consumed cost. Note that without the top-$k$ strategy, RefQD will use all the levels of DDA to re-evaluate. As shown in Figure~\ref{fig-top-k-reeval}, using a smaller $k$ will generally achieve a satisfactory performance, and we use $k=1$ in our experiments.

\textbf{Ablation studies.} We finally perform ablation studies on RefQD to analyze the effectiveness of each component, such as periodic re-evaluation, maintaining a DDA, top-$k$ re-evaluation, and learning rate decay for the representation part. The experimental results demonstrate their significance in enhancing the resource efficiency. Due to space limitation, detailed results can be found in Appendix~\ref{app:ablation-studies}.

\section{Conclusion}

In this paper, we first emphasize the importance of resource efficiency during the training phase of QD algorithms and propose RefQD, a novel and effective method to enhance the resource efficiency of QD. Experimental results on the QDax suite and Atari environments with limited resources demonstrate the effectiveness of RefQD, which utilizes only 3.7\% to 16\% GPU memories of the well-known PGA-ME, and achieves comparable or even superior performance, with nearly the same wall-clock time and number of samples.

RefQD improves the resource efficiency of QD by considering the properties and principles of the QD algorithm itself, i.e., decomposing a neural network into a large representation part and a small decision part, and sharing the representation part with all decision parts. It is a general framework, which can be equipped with advanced sample-efficient operators to further improve the performance, and can also be combined with other general approaches such as model pruning to further decrease the computational overhead. The deep decision archive of RefQD can also be distilled into a single decision policy by archive distillation methods to reduce costs during the deployment phase. We hope this work can encourage the application of QD in more complex and challenging scenarios, such as embodied robots and large language models.
One limitation of this paper is that we only examine the effectiveness of RefQD through empirical studies, without giving a mathematical theoretical analysis~\cite{qd-theory}.

\section*{Impact Statement}

This paper presents work whose goal is to advance the field of quality-diversity optimization. Our work enables more efficient use of computational resources in QD algorithms and applications, making it possible to solve large-scale quality-diversity optimization problems with limited resources.

\section*{Acknowledgements}

The authors want to thank the anonymous reviewers for their helpful comments and suggestions. This work was supported by the National Science and Technology Major Project (2022ZD0116600) and National Science Foundation of China (62276124).

\bibliography{main}
\bibliographystyle{icml2024}

\newpage
\appendix
\onecolumn

\section{Detailed Settings}\label{app:detailed-settings}

\subsection{Algorithms}

For a fair comparison, we unify the common hyperparameters of these methods on all the eight environments. The other hyperparameters of each method are set as the corresponding original paper. 

The network structure on QDax is:
\begin{align*}
        \text{state} &\to \textbf{MLP}(256)\to \textbf{tanh} \to \textbf{MLP}(256) \to \textbf{tanh}\to \textbf{MLP}(\text{number of actions}) \to \textbf{tanh} \to \text{action}
\end{align*}
The CNN structure on Atari is:
\begin{align*}
        \text{state} &\to \textbf{CNN}(8 \times 8 \times 32)\to \textbf{leaky\_relu}
        \to \textbf{CNN}(4 \times 4 \times 64)\to \textbf{leaky\_relu}
        \to \textbf{CNN}(3 \times 3 \times 64)\to \textbf{leaky\_relu}\\
        &\to \textbf{MLP}(512)\to \textbf{leaky\_relu}\to \textbf{MLP}(\text{number of actions}) \to \text{action}
\end{align*}
The ResNet network structure on Atari is:
\begin{equation*}
        \text{state} \to \textbf{ResNet34}
        \to \textbf{MLP}(512)\to \textbf{leaky\_relu}\to \textbf{MLP}(\text{number of actions}) \to \text{action}
\end{equation*}

Then, we introduce the two types of variation operators used in our experiments.

\textbf{IsoLineDD.} IsoLineDD~\citep{arm-gecco} is a popular evolutionary operator used in several QD algorithms~\citep{PGA-ME,uncertain-qd-arxiv,NeuroevoSkillDiscovery,accelerated-qd}.
Considering two parent solutions $\boldsymbol x_1$ and $\boldsymbol x_2$, the offspring solution $\boldsymbol x'$ generated by the IsoLineDD operator is sampled as follows:
\begin{equation*}
    \boldsymbol x' = \boldsymbol x_1 + \sigma_1 \cdot \mathcal{N} (\boldsymbol 0, \mathbf I) + \sigma_2 \cdot (\boldsymbol x_2 - \boldsymbol x_1) \cdot \mathcal{N} (0, 1),
\end{equation*}
where $\sigma_1 = 0.005$ and $\sigma_2 = 0.05$ in this paper, $\mathbf I$ denotes the identity matrix, $\mathcal{N} (0, 1)$ and $\mathcal{N} (\boldsymbol 0, \mathbf I)$ denote a random number and a random vector sampled from the standard Gaussian distribution, respectively. 

\textbf{Policy gradient operators.}
The policy gradient operator maintains a critic and a greedy actor in many QD algorithms~\citep{PGA-ME,qd-pg,DQD-RL,NeuroevoSkillDiscovery,accelerated-qd}, and we adopt this setting as well. At the start of each generation, the critic is trained with TD loss, while the greedy actor is simultaneously trained with policy gradient. Subsequently, the parent solutions are updated with policy gradient, using the critic in the variation process. We employ the policy gradient method TD3, whose hyperparameters are presented in Table~\ref{table:td3}.

\begin{table}[htpb]
\caption{The hyperparameters of TD3.}
\centering
\vspace{0.3em}
\begin{tabular}{lc}
\toprule
Hyperparameter & Value \\ \midrule
Critic hidden layer size & $[256, 256]$ \\
Policy learning rate & $1\times10^{-3}$ \\
Critic learning rate & $3\times10^{-4}$ \\
Replay buffer size & $1\times10^6$ \\
Training batch size & $32$ \\
Policy training steps & $100$ \\
Critic training steps & $300$ \\
Reward scaling & $1.0$ \\
Discount & $0.99$ \\
Policy noise & $0.2$ \\
Policy clip & $0.5$ \\ \bottomrule
\end{tabular}
\label{table:td3}
\end{table}

\textbf{Implementation of RefQD.} We provide an implementation of RefQD using uniform parent selection and reproduction operators of PGA-ME in the experiment. The greedy actor of the reproduction operator is also decomposed into two parts, and the representation part of the greedy actor is also shared with other policies. To train the shared representation part, it is combined with the decision part of the greedy actor as well as several decision parts sampled from the decision archive to obtain the policy gradient. To train the decision part of the greedy actor and to variate the decision parts of the parent solutions, they are combined with the shared representation part to take actions.

\newpage

\textbf{Detailed settings of methods.} The settings of the methods are summarized as follows.
\begin{itemize}
    \item \textbf{PGA-ME}~\citep{PGA-ME,pga-me-telo} uses the IsoLineDD operator, and the policy gradient in the variation process. The proportion of the offspring solutions generated by the two operators is both $0.5$, which is the same as~\citep{PGA-ME,accelerated-qd}.
    \item \textbf{PGA-ME (s)} is the same as PGA-ME, except that the population size is set to $8$ to make it have a similar GPU memories usage to RefQD.
    \item \textbf{DQN-ME} uses the DQN as the policy to handle the discrete action space in Atari.
    \item \textbf{RefQD} uses the variation operator as same as PGA-ME. The number of levels of the DDA is $4$, and the top level is re-evaluated every $50$ iterations.
\end{itemize}

\subsection{Computational Resources}\label{app:B3-resources}
All the experiments are conducted on an NVIDIA RTX 3090 GPU (24~GB) with an AMD Ryzen 9 3950X CPU (16~Cores).

\section{Additional Results}\label{app:additional-results}

\subsection{More Complex Policy Architectures} \label{app:resnet}

\begin{figure*}[htbp]\centering
\includegraphics[width=0.48\linewidth]{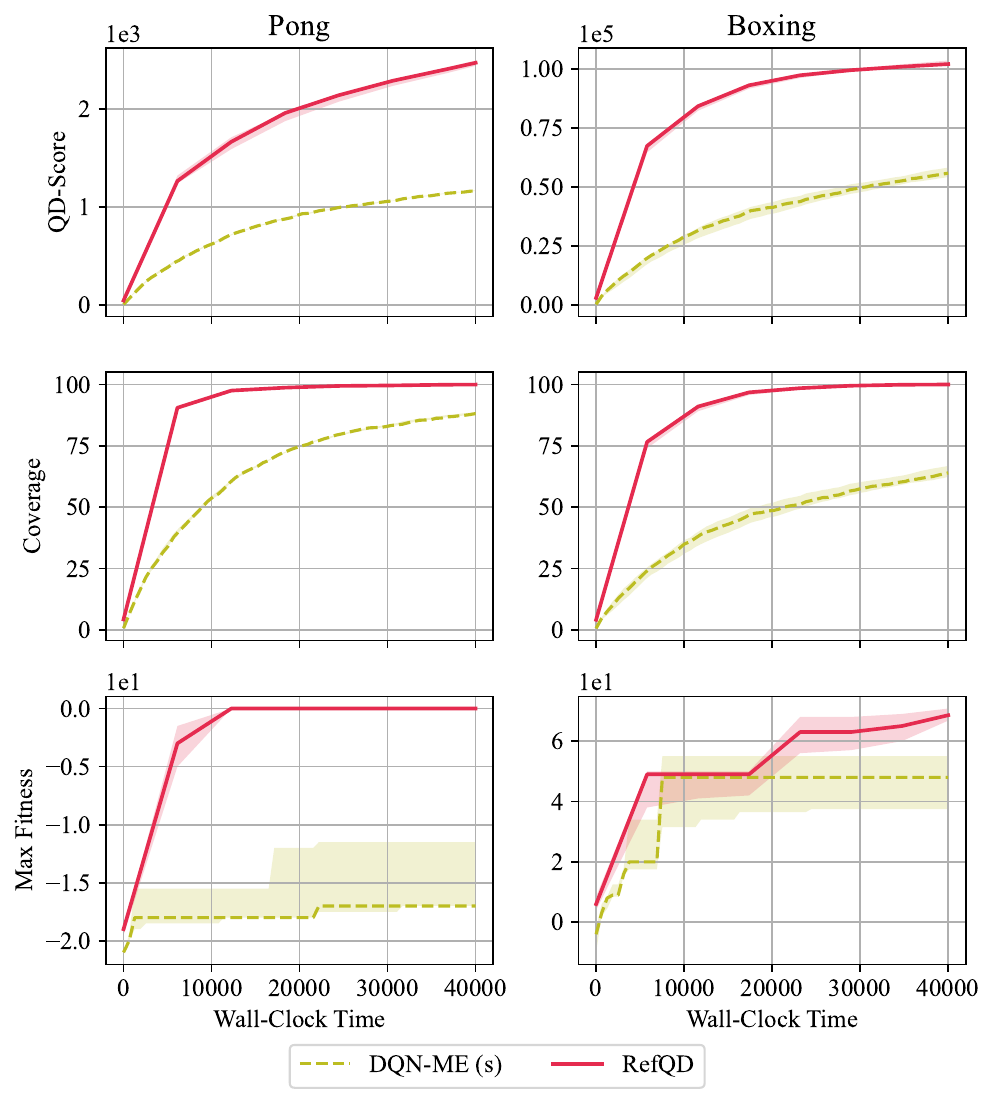}
\caption{Performance comparison in terms of QD-Score, Coverage, and Max Fitness using ResNet as policy networks on two environments of Atari. The medians and the first and third quartile intervals are depicted with curves and shaded areas, respectively.}\label{fig-resnet}
\end{figure*}

To examine the scalability of RefQD in challenging tasks with higher-dimensional decision space, we also conduct experiments in the Atari environments with ResNet policy architecture. Due to the high-dimensional decision spaces, the default settings of DQN-ME run out of memory and fail to work. Thus, we compare our method with DQN-ME (s), which has a small number of offspring solutions, using smaller GPU memory than the default settings but still requiring to maintain the whole archive in unlimited RAM. As shown in Figure~\ref{fig-resnet}, RefQD with limited resources achieves significantly better performance than DQN-ME (s).

\subsection{Decomposition Strategies} \label{app:decomp}

\begin{figure*}[htbp]\centering
\includegraphics[width=0.96\linewidth]{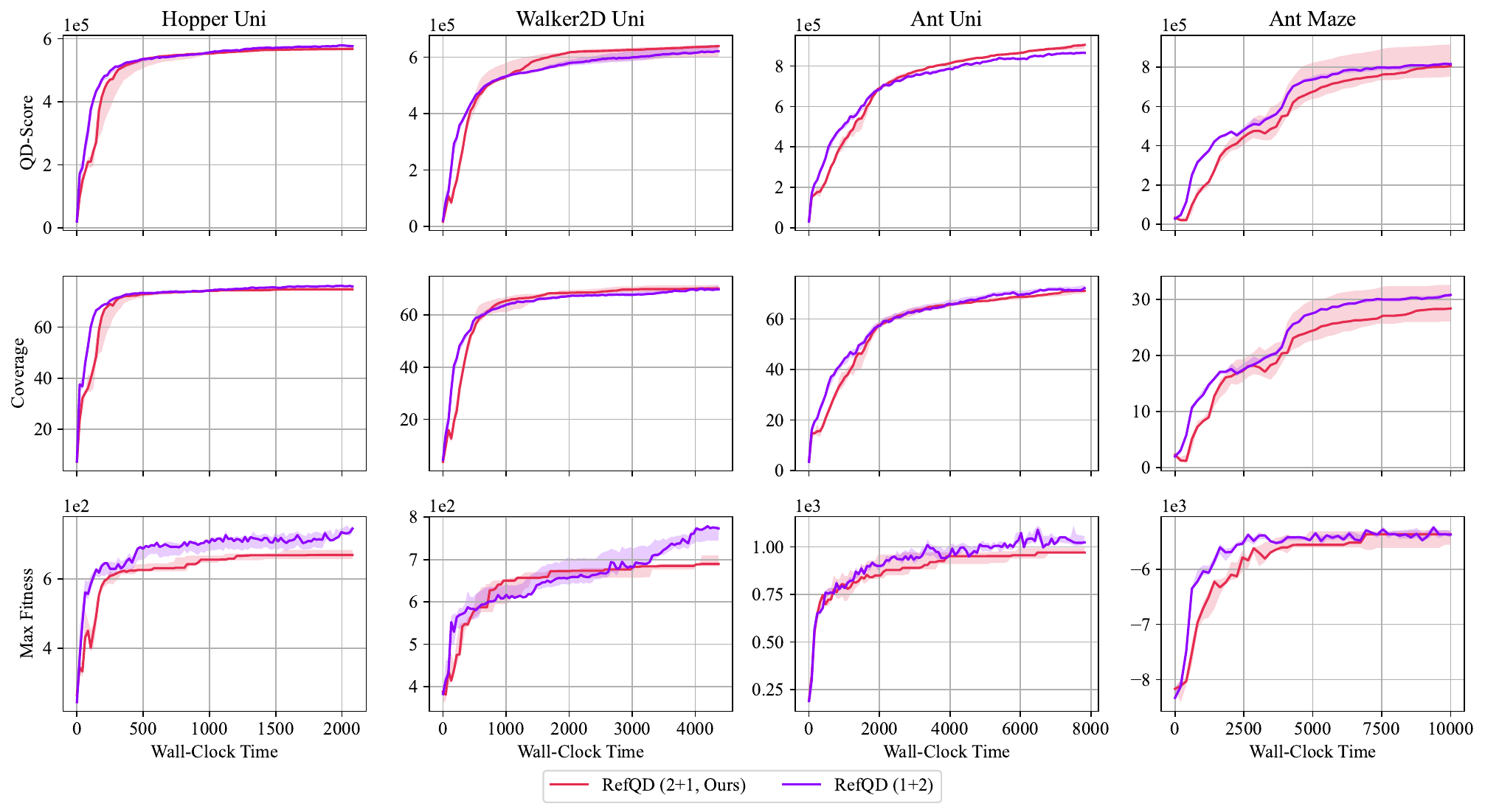}
\caption{Performance comparison with different decomposition strategies in terms of QD-Score, Coverage, and Max Fitness. The medians and the first and third quartile intervals are depicted with curves and shaded areas, respectively.}\label{fig-decomp}
\end{figure*}

The decomposition strategy is a crucial setting as it significantly impacts resource utilization. For example, when computational resources are very limited, the representation part should be shared as much as possible to reduce overhead, just as we do in the default settings in our paper. Generally, the more we share, the less computational resources are required, but the greater potential for performance drop. We conduct experiments to analyze the influence of the decomposition strategies. We compare the performance of (2+1) and (1+2), where (2+1) denotes that the representation part and decision part have 2 and 1 layers, respectively, and (1+2) denotes that they have have 1 and 2 layers, respectively. As shown in Figure~\ref{fig-decomp}, when sharing less representation, RefQD has a better Max Fitness, due to the stronger ability from additional resource overhead. In practical use, the decomposition strategy of RefQD can be set according to the available computational resources. It can also be adaptive based on the task or network architecture.

\subsection{Ablation Studies} \label{app:ablation-studies}

We also consider the following variants of RefQD:
\begin{itemize}
 \item \textbf{RefQD w/o TR}: The same as RefQD, except that it does not use top-$k$ re-evaluation (TR), but re-evaluates the whole archive.
    \item \textbf{RefQD w/o TR \& LRD}:
    The same as RefQD w/o TR, except that it does not use the learning rate decay (LRD) of the representation part.
    \item \textbf{Vanilla-RefQD w/ Re-SS} (i.e., RefQD w/o TR, LRD, \& DDA):
    The same as Vanilla-RefQD, except that it performs the re-survivor selection (Re-SS).
\end{itemize}

\begin{figure*}[htbp]\centering
\includegraphics[width=0.96\linewidth]{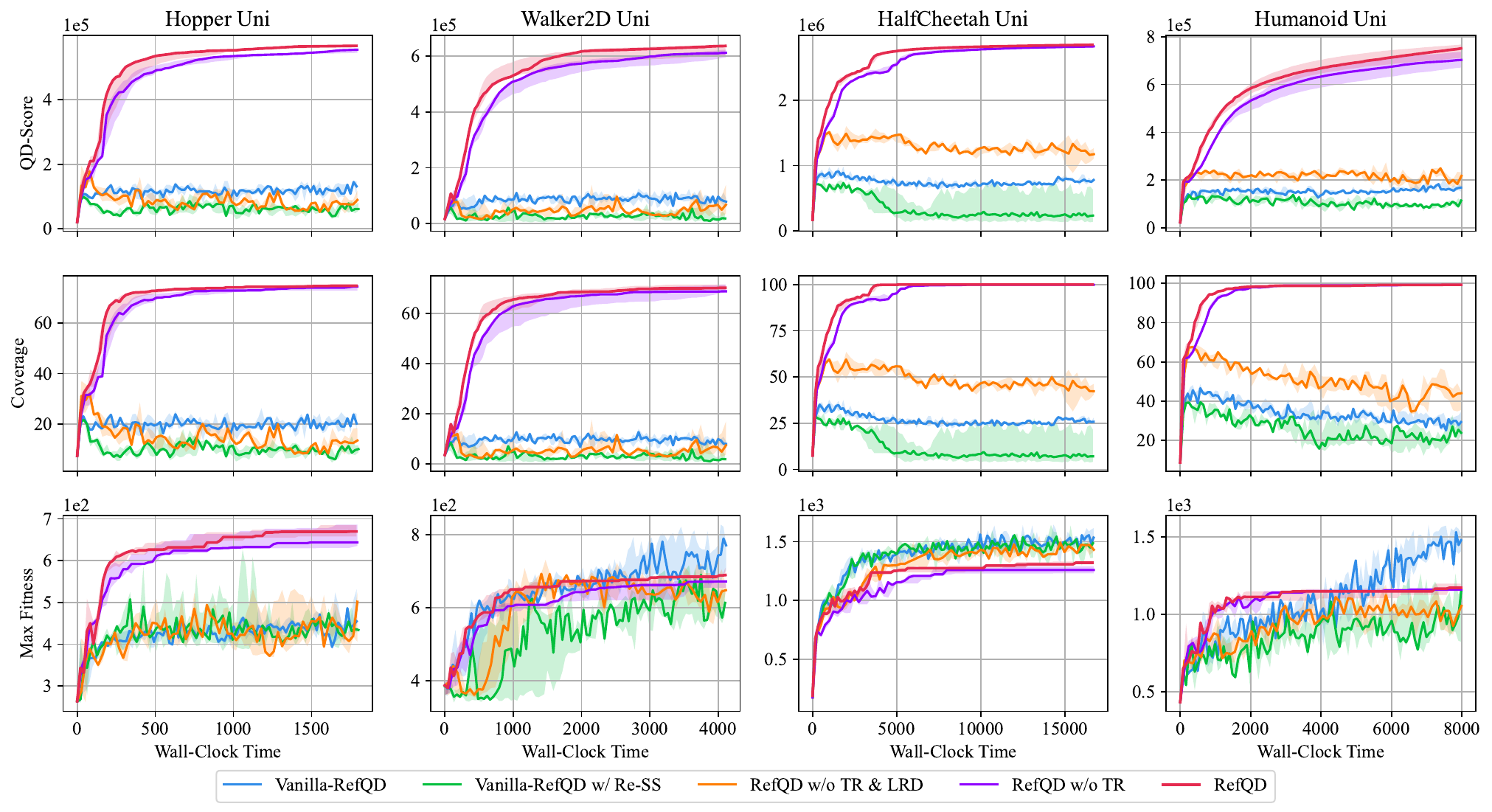}\\
\includegraphics[width=0.96\linewidth]{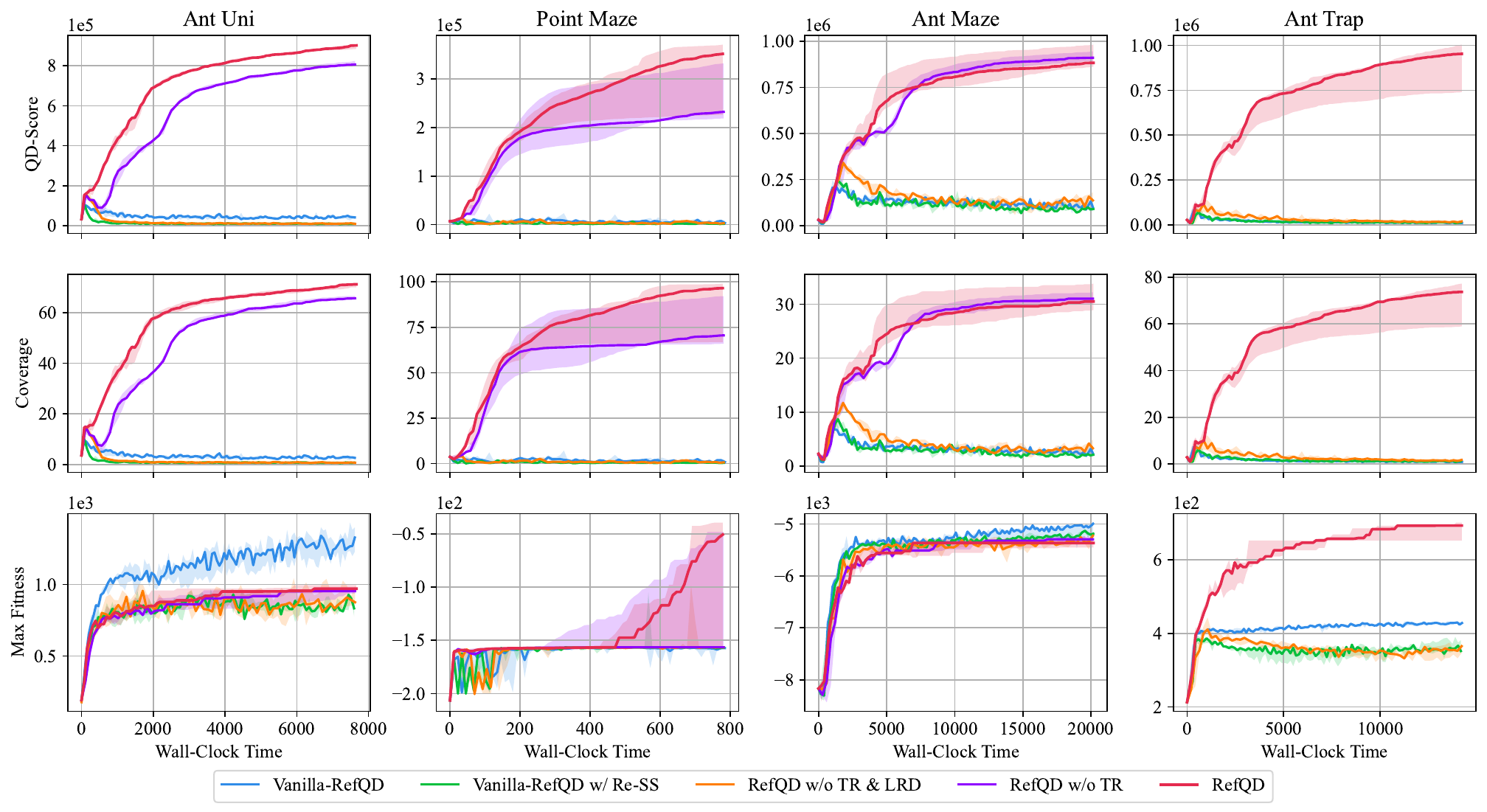}
\caption{Performance comparison in terms of QD-Score, Coverage, and Max Fitness in ablation studies. The medians and the first and third quartile intervals are depicted with curves and shaded areas, respectively.}\label{fig-ablation}
\end{figure*}

As shown in Figure~\ref{fig-ablation}, Vanilla-RefQD has the serious mismatch issue and does not perform well. Vanilla-RefQD w/ Re-SS performs even worse, as it may waste a lot of learned knowledge in the decision parts. RefQD performs the best, and removing the top-$k$ re-evaluation or the learning rate decay for the representation part results in a performance degradation.

\subsection{Time-steps vs. QD metrics}

\begin{figure*}[!hb]\centering
\includegraphics[width=0.96\linewidth]{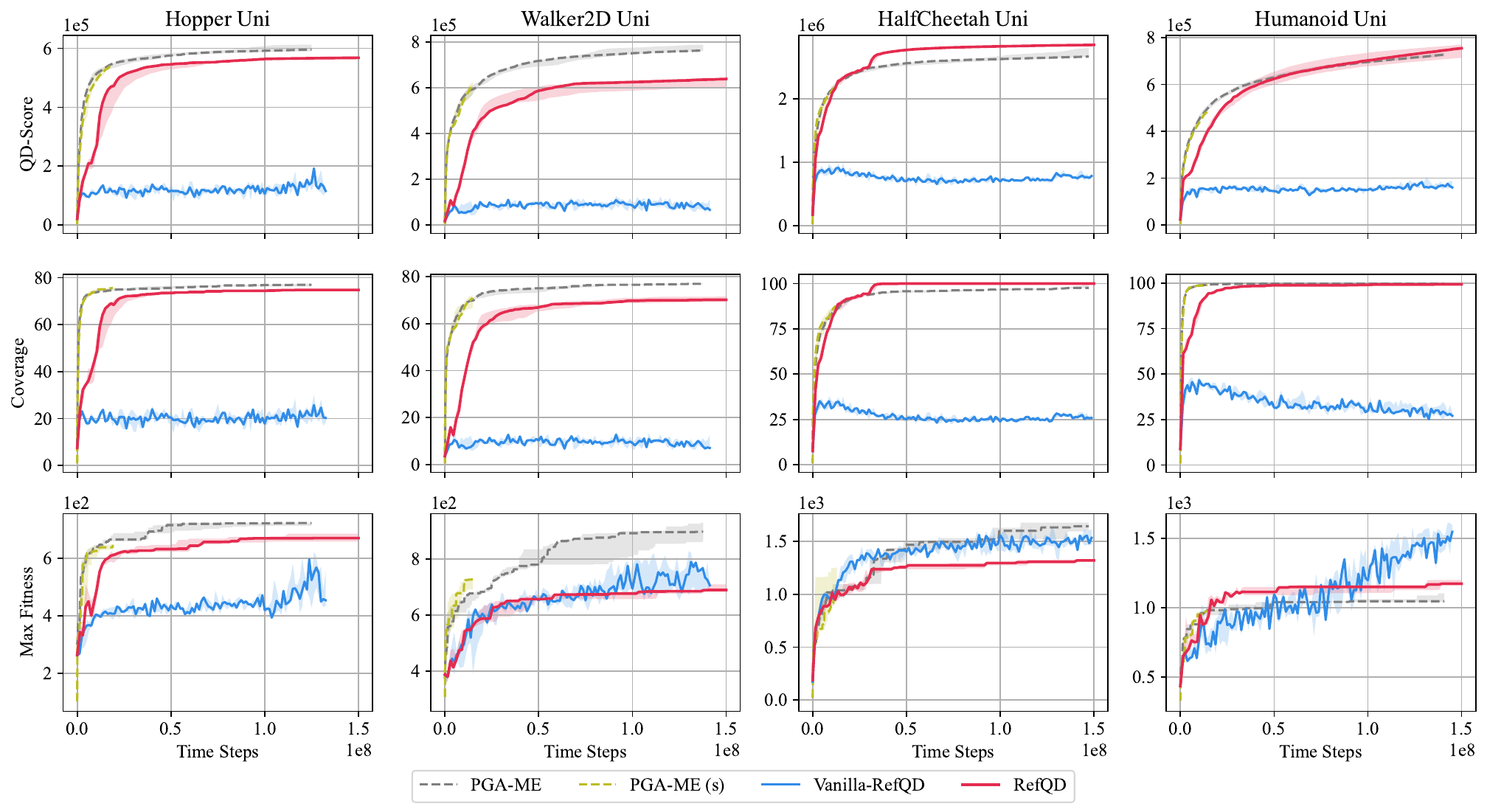}\\
\includegraphics[width=0.96\linewidth]{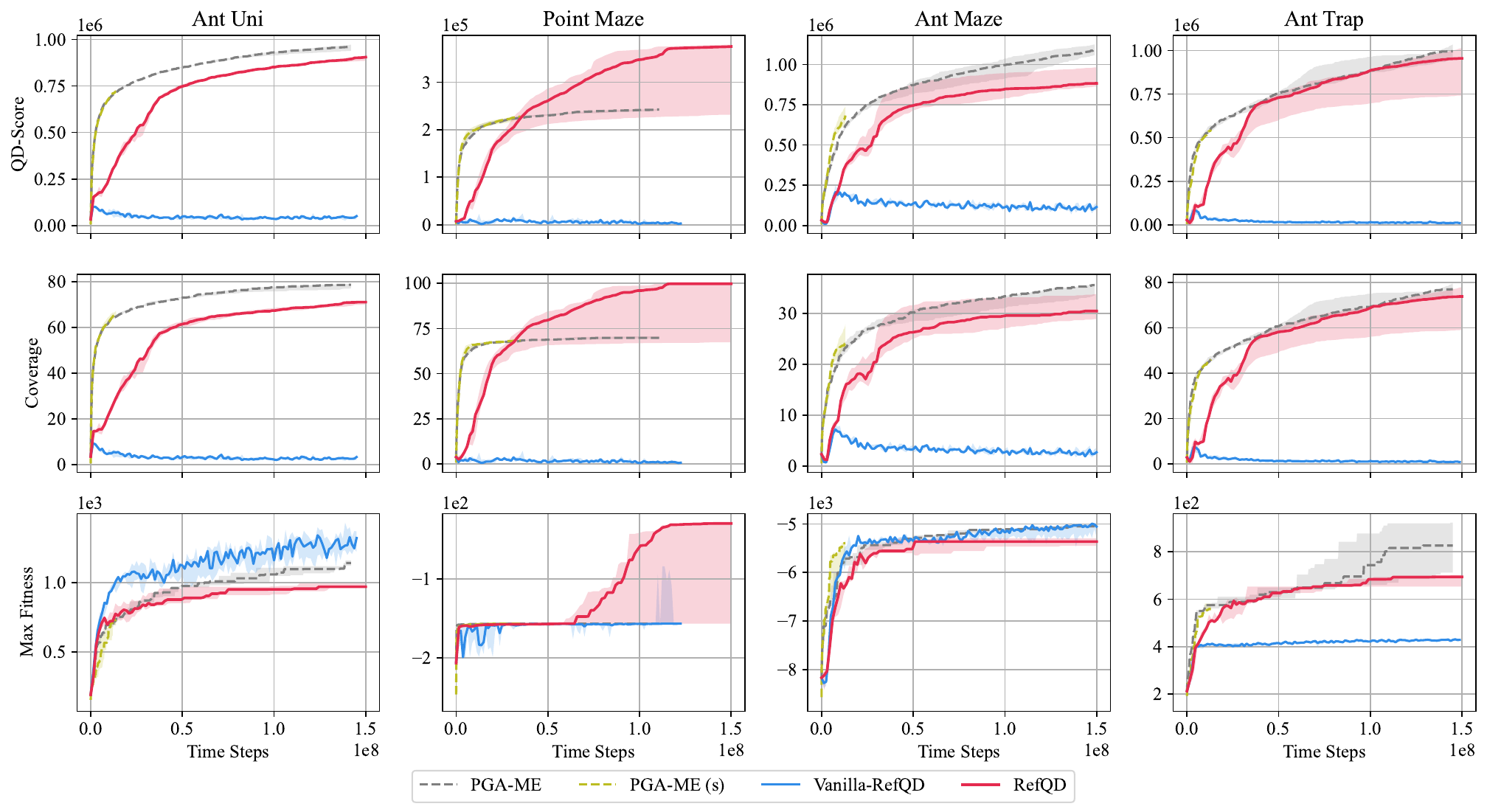}
\caption{Performance comparison in terms of QD-Score, Coverage, and Max Fitness w.r.t. the number of time-steps. The medians and the first and third quartile intervals are depicted with curves and shaded areas, respectively.}\label{fig-samples}
\end{figure*}

Apart from the wall-clock time, we also compare the sample efficiency of the methods. As shown in Figure~\ref{fig-samples}, RefQD with limited resources also achieves similar sample efficiency to PGA-ME and PGA-ME (s), and performs even better in HalfCheetah Uni, Humanoid Uni, and Point Maze.

\subsection{Combined with EDO-CS}

\begin{figure*}[htbp]\centering
\includegraphics[width=0.96\linewidth]{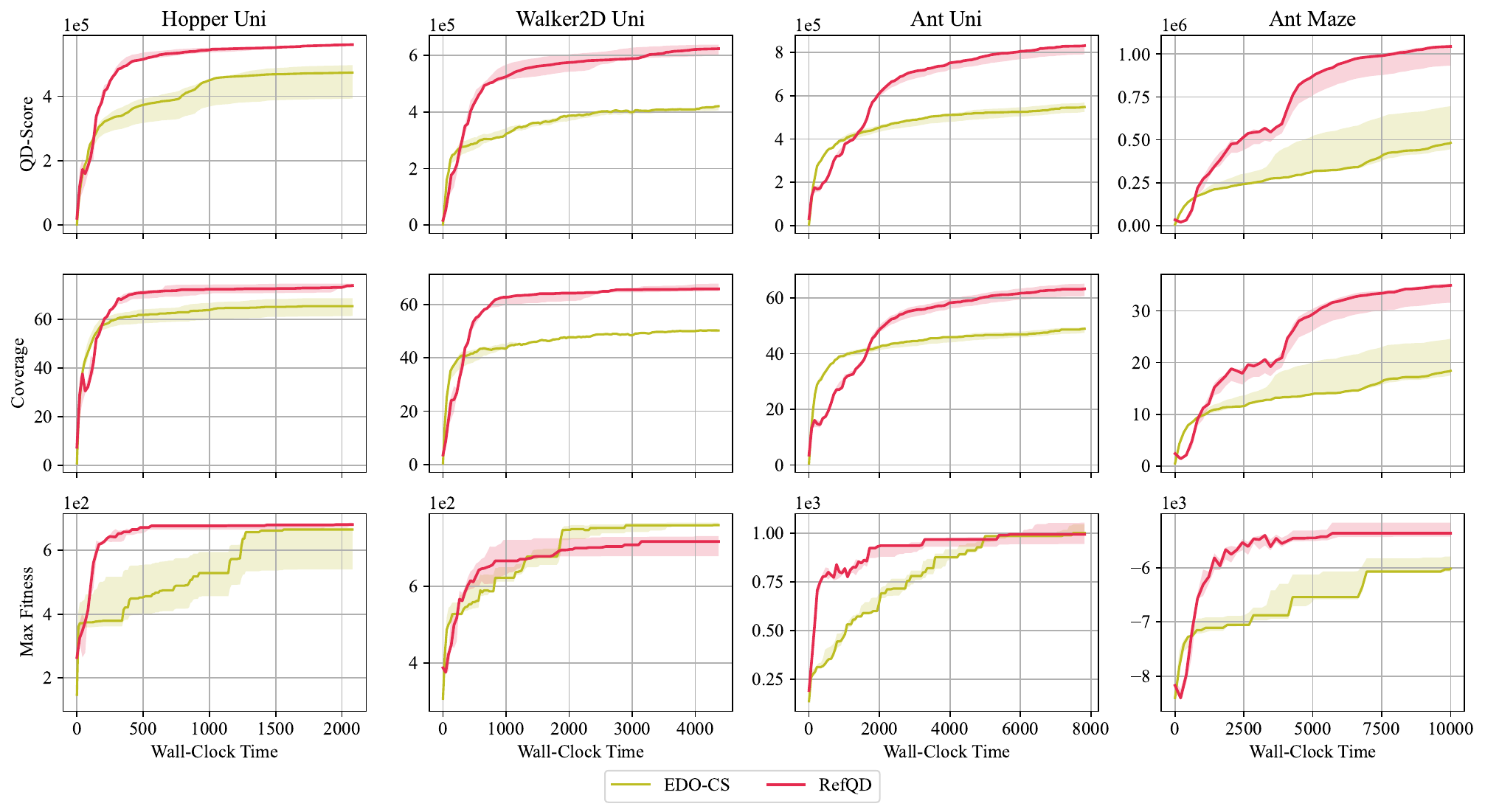}
\caption{Performance comparison of EDO-CS-based methods in terms of QD-Score, Coverage, and Max Fitness. The medians and the first and third quartile intervals are depicted with curves and shaded areas, respectively.}\label{fig-edocs}
\end{figure*}

RefQD is a general method that can be integrated with different QD algorithms. We also adopt EDO-CS~\cite{EDO-CS} as another QD algorithm. As shown in Figure~\ref{fig-edocs}, the EDO-CS variant of RefQD with limited resources also achieves similar or better performance than the original version of EDO-CS, demostrating the generalization of RefQD.

\end{document}